\documentclass[letterpaper, 10 pt, conference]{ieeeconf} 
\IEEEoverridecommandlockouts % for the thanks block
\usepackage{graphicx} % for pdf, bitmapped graphics files
\graphicspath{{./figs}}
\usepackage{times}
\usepackage{amsmath}
\usepackage{amssymb}
\usepackage{dsfont}
\usepackage{cite}
\usepackage{url}
\usepackage[caption=false,font=footnotesize]{subfig}
\usepackage{gensymb}
\usepackage{booktabs}
\usepackage{multirow}
\usepackage[T1]{fontenc}
\usepackage{flushend}

\title{\LARGE \bf Towards Efficient Roadside LiDAR Deployment: A Fast Surrogate Metric Based on Entropy-Guided Visibility}
\author{Yuze Jiang$^1$, Ehsan Javanmardi$^1$, Manabu Tsukada$^1$, and Hiroshi Esaki$^1$
\thanks{$^1$Graduate School of Information Science and Technology, The University of Tokyo, Tokyo, 113-0033, Japan. (E-mail: \{uiryuu,ejavanmardi,mtsukada\}@g.ecc.u-tokyo.ac.jp, hiroshi@wide.ad.jp)}}
\begin{document}
\maketitle
\thispagestyle{empty}
\pagestyle{empty}

\begin{abstract}
The deployment of roadside LiDAR sensors plays a crucial role in the development of Cooperative Intelligent Transport Systems (C-ITS). However, the high cost of LiDAR sensors necessitates efficient placement strategies to maximize detection performance. Traditional roadside LiDAR deployment methods rely on expert insight, making them time-consuming.  Automating this process, however, demands extensive computation, as it requires not only visibility evaluation but also assessing detection performance across different LiDAR placements. To address this challenge, we propose a fast surrogate metric, the Entropy-Guided Visibility Score (EGVS), based on information gain to evaluate object detection performance in roadside LiDAR configurations. EGVS leverages Traffic Probabilistic Occupancy Grids (TPOG) to prioritize critical areas and employs entropy-based calculations to quantify the information captured by LiDAR beams. This eliminates the need for direct detection performance evaluation, which typically requires extensive labeling and computational resources. By integrating EGVS into the optimization process, we significantly accelerate the search for optimal LiDAR configurations. Experimental results using the AWSIM simulator demonstrate that EGVS strongly correlates with Average Precision (AP) scores and effectively predicts object detection performance. This approach offers a computationally efficient solution for roadside LiDAR deployment, facilitating scalable smart infrastructure development.
\end{abstract}

\section{Introduction}
Smart roadside infrastructure plays a critical role in modern Cooperative Intelligent Transport Systems (C-ITS), enabling accurate traffic monitoring~\cite{Zhang2020-fs,Zhao2019-jq}, cooperative perception~\cite{Sun2022-ry,Gao2024-fv}, and localization~\cite{Jiang2024-ea}. Smart infrastructure usually comes with expensive high-end sensor sets, with light detection and ranging (LiDAR) sensors being one of the most important, as they can provide highly accurate depth data of perceived objects. This results in more accurate positional data and better traffic understanding for different types of road users.

As learning-based 3D object detection methods utilizing point cloud data continue to advance in accuracy and efficiency, LiDARs are now capable of detecting objects with greater precision and lower computational requirements. However, the influence of LiDAR placement on model performance is often overlooked, despite its significant impact on the quality of input data and, consequently, object detection accuracy. This is particularly critical for roadside LiDARs, where variations in placement can lead to vastly different detection results. These discrepancies are primarily driven by two factors: 1) the extent to which the LiDAR effectively covers high-traffic areas and 2) the occlusions caused by stationary objects within the environment. Despite the importance of these factors, a comprehensive roadside LiDAR performance metric that accounts for them remains absent.

From an economic perspective, the high cost of LiDAR sensors makes it challenging for governments to deploy them extensively across critical sections of a city~\cite{Gao2018-fy}. Therefore, it is essential to minimize the number of sensors while maximizing the coverage area and object detection capabilities. However, evaluating the object detection performance of a specific LiDAR configuration in the real world is a highly time-consuming process~\cite{Zhang2024-wq}. Existing studies usually set up experiments in controlled environments and use Real-Time Kinematic (RTK)-GPS to collect vehicle positions for ground truth. Then, roadside LiDAR is deployed to collect raw data, and learning-based object detection methods are used to detect vehicles. The perception performance is then evaluated by comparing the object detection results with the ground truth. Alternative methods include setting up a traffic simulator using a game engine, collecting point cloud traffic data, training 3D object detection models on the collected data, and then assessing their performance for each roadside LiDAR configuration~\cite{Ma2024-av}. To efficiently identify optimal LiDAR configurations, there is a pressing need for a method to quickly compare the performance of different configurations without relying on simulations or full-scale 3D object detection evaluations, thereby accelerating the search process.

\begin{figure*}
    \centering
    \includegraphics[width=\linewidth]{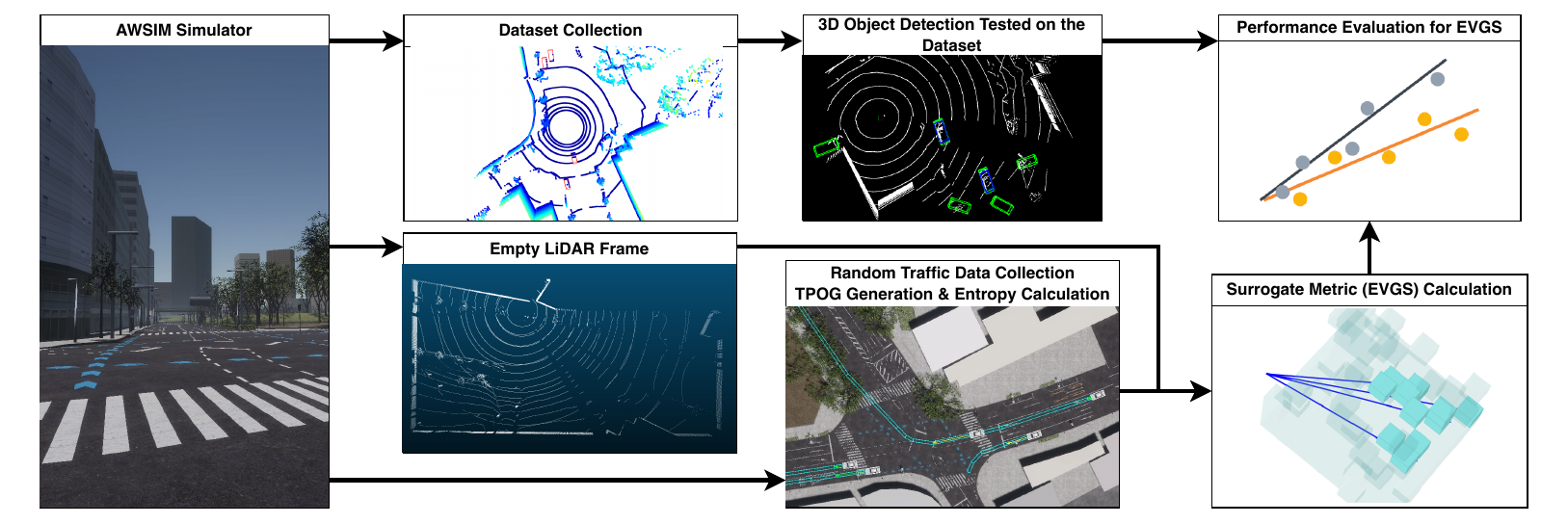}
    \caption{Overview of the proposed surrogate metric and its performance evaluation. Random traffic data collection is needed only once for each scenario for different roadside LiDAR placements inside the ROI.}
    \label{fig:overview}
\end{figure*}

In this paper, we propose the Entropy-Guided Visibility Score (EGVS) for evaluating the performance of roadside LiDAR in 3D object detection, designed to be computationally efficient. As shown in Figure~\ref{fig:overview}, we conducted experiments to validate our surrogate metric in a simulation environment. We collected traffic data from the simulator and used empty LiDAR frames to calculate the EGVS via ray tracing. We then created datasets for each configuration and ran 3D object detection models on them to gather the ground truth. We demonstrate that our surrogate metric strongly correlates with the 3D object detection precision of roadside LiDAR configurations.

The main contributions of this paper can be summarized as follows:

\begin{itemize}
    \item We employed an entropy-based method to evaluate various roadside LiDAR placements in a realistic environment modeled after real-world settings, demonstrating that our approach is applicable to real-world LiDAR deployment scenarios.
    \item The proposed EGVS exhibits a strong correlation with actual perception performance. As a result, decision-makers can rely on the surrogate metric to compare different roadside LiDAR configurations without needing to obtain the actual perception performance.
    \item Our proposed metric takes both occlusions caused by static objects and traffic patterns into account, providing a more reliable object detection performance comparison among different roadside LiDAR placements.
\end{itemize}

\section{Related Work}
\subsection{Smart Infrastructure and C-ITS}
In the context of C-ITS, roadside infrastructure plays a critical role in facilitating real-time data exchange between vehicles and infrastructure. By bridging the gap between vehicles and their surrounding environment, smart infrastructure contributes to safer, more efficient, and sustainable transportation systems~\cite{Han2024-ic}. This infrastructure is equipped with sensors, communication devices, and computing units that collaborate with connected vehicles. The sensors installed on smart infrastructure are capable of monitoring various road users~\cite{Cui2019-yl,El-Hamdani2020-ee,Hernandez-Jayo2016-hn}, detecting obstacles~\cite{Ding2023-ev}, and measuring traffic density~\cite{Datondji2016-mu}. Among these, 3D object detection has become a fundamental component in enabling these functionalities.

However, assessing the effectiveness of sensor placement is a complex challenge. While most learning-based object detection models use Average Precision (AP) as a standard metric to evaluate detection performance, directly calculating AP for different sensor placements to determine the optimal configuration is economically unfeasible for roadside sensors. Computing AP requires both ground truth labels and prediction data from the sensors to create datasets for each potential sensor placement. Collecting ground truth data in real-world scenarios, however, is highly challenging and resource-intensive.

\subsection{Sensor Placement for Autonomous Vehicles}
Existing research on the sensor placement problem primarily focuses on optimizing the placement of vehicle sensors. Many studies utilize entropy-based methods to evaluate different LiDAR configurations. For instance, Roos~\textit{et al.}~\cite{Roos2021-lb} proposed an entropy-based score to represent the perception performance of point cloud-based vehicle sensors. However, in their evaluation process, they relied solely on recall as the performance indicator. While recall is crucial for assessing object detection coverage, it overlooks false positives, which are equally important for evaluating the quality of predictions in sensor placement.

Hu~\textit{et al.}~\cite{Hu2022-rz} introduced the entropy-based surrogate metric S-MIG, which uses a probabilistic occupancy grid to evaluate vehicle LiDAR placement. However, their results do not exhibit a strong correlation between S-MIG and AP values, as the trends are inconsistent and non-monotonic. A reliable surrogate metric should demonstrate a predictable relationship, where AP increases or decreases consistently with changes in S-MIG. The observed fluctuations in their results cast doubt on the validity of S-MIG as a dependable surrogate for performance. More recently, Li~\textit{et al.}~\cite{Li2024-uy} proposed a more advanced approach using semantic occupancy grids to calculate entropy-based surrogate scores as indicators of perception performance. Their results show a stronger correlation between the surrogate metric and AP values compared to S-MIG, indicating an improvement in predictive reliability.

\subsection{Roadside LiDAR Deployment Optimization}
While the placement of vehicle sensors has been relatively well studied in the literature, the question of optimal roadside sensor placement largely remains unanswered. Unlike vehicle sensors, roadside sensors offer greater flexibility and a wider range of placement options. However, the performance of roadside LiDAR object detection is significantly influenced by the presence of blind spots. Optimizing sensor placement solely based on eliminating blind spots, such as in~\cite{Kim2023-ry}, however, does not necessarily guarantee improved object detection performance.

The authors in~\cite{He2024-ke} proposed a method designed to automatically adjust the beam configuration and height of a roadside LiDAR to maximize the perception range for vehicles. This approach leverages the assumption that future LiDAR technologies will include beam-configurable capabilities, allowing for dynamic adjustments to improve detection. However, the work does not address how the physical placement of the roadside LiDAR itself might impact overall object detection performance, leaving this critical aspect unexplored. In contrast, the work in~\cite{Jiang2023-om} proposed a greedy algorithm-based method to optimize roadside LiDAR placement. Jiang~\textit{et al.} used a network-based approach to predict object detection precision using only empty LiDAR scans from specific placements. However, this method relies solely on point cloud data and does not account for traffic conditions. Similarly, Jin \textit{et al.}~\cite{Jin2022-ti} introduced an entropy-based approach to evaluate roadside LiDAR placements, but it overlooked the spatial position of the LiDAR and its relationship to realistic traffic patterns. In contrast, our method improves upon these approaches by utilizing ray tracing algorithms to identify intersecting voxels from LiDAR beams and evaluate the information contained within them from the traffic data. This allows for a more accurate assessment of LiDAR placements by incorporating interactions with traffic.

\section{Methodology}

\subsection{Problem Formulation}
Our work aims to develop a fast and reliable evaluation method for assessing object detection performance in roadside LiDAR deployments. Specifically, we propose a surrogate metric that quantifies the effectiveness of a roadside LiDAR at a given configuration. This metric enables decision-makers in road infrastructure planning to efficiently compare different LiDAR placement options without the need to compute the actual perception precision, which requires annotated datasets and significant computational resources. Unlike most existing approaches, which often neglect the uneven nature of traffic when optimizing LiDAR placement, our proposed metric incorporates traffic data alongside static object occlusions to provide a more accurate estimation of a LiDAR’s performance for a given configuration.

To evaluate the object detection performance of roadside LiDARs, we first define a Region of Interest (ROI). For simplicity, the ROI is modeled as a rectangular cuboid characterized by its center point $[x, y, z]$ and dimensions $[w, l, h]$. Following the methods outlined in \cite{Hu2022-rz} and \cite{Li2024-uy}, we discretize the ROI into a set of voxels with a fixed resolution $\delta$, resulting in the following voxel set:

\begin{equation}
    \mathcal{V} = \{v_1, v_2, \cdots, v_M\}, M = \frac{w}{\delta} + \frac{l}{\delta} + \frac{h}{\delta}.
\end{equation}

The configuration of a single roadside LiDAR is defined as:

\begin{equation}
    C = [x, y, z],
\end{equation}

where $x, y$ represents the horizontal position of the LiDAR, and $z$ specifies its mounting height.

Our optimization goal is to maximize the AP of the deployed roadside LiDARs within the prescribed ROI. AP is a widely used metric in object detection tasks, as it balances precision and recall, making it suitable for evaluating the effectiveness of LiDAR placements in detecting objects. By restricting the evaluation to the ROI, we ensure that objects or vehicles outside of the region do not affect the computed AP value. This allows us to focus solely on the detection performance within the intended area of coverage. The traffic data is modeled as a probabilistic distribution of moving objects within the ROI, and our surrogate metric accounts for their occlusions and detectability.

\begin{figure}[h]
    \centering
    \includegraphics[width=0.9\linewidth]{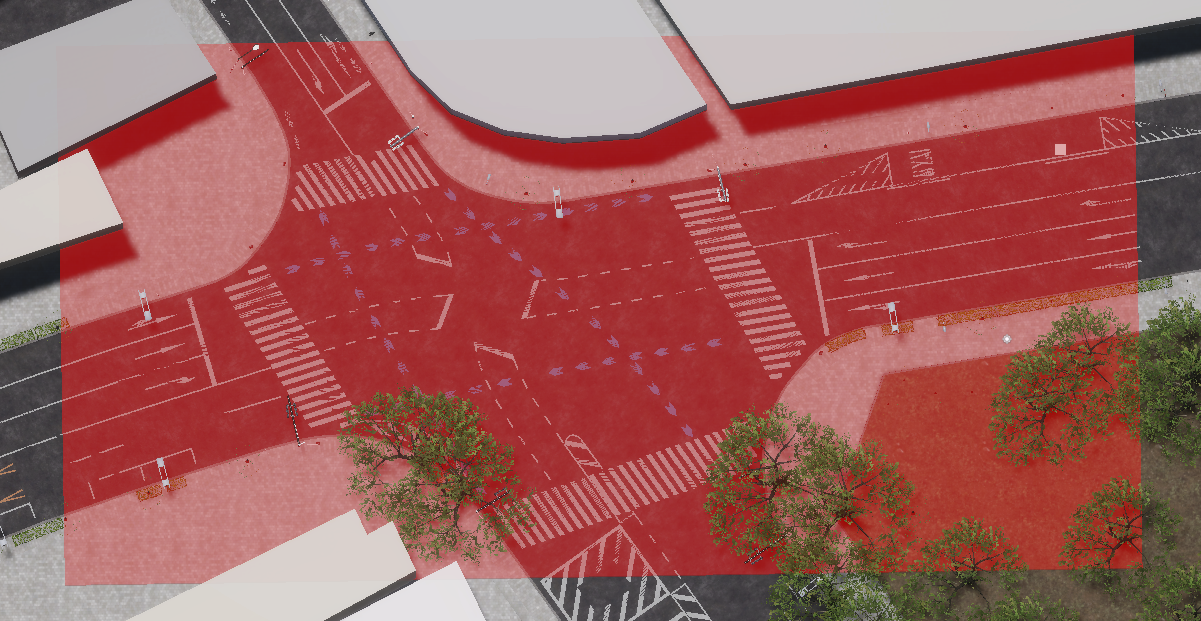}
    \caption{The testing environment in the Nishishinjuku map, which is based on the real landscape in Shinjuku Ward, Tokyo. The red box denotes the region of interest (ROI) with the dimensions of $50\,\mathrm{m} \times 100\,\mathrm{m} \times 5\,\mathrm{m}$.}
    \label{fig:roi}
\end{figure}

\subsection{Traffic Data Collection}

We utilized AWSIM~\cite{TIER-IV-incUnknown-pv} to simulate traffic in a realistically simulated environment based on Shinjuku Ward, Tokyo, Japan. AWSIM is a Unity-based driving simulator capable of loading real-world maps and road lane structures. It integrates RobotecGPULiDAR (RGL)~\cite{AiUnknown-cu}, a high-performance LiDAR simulation library, which provides accurate and efficient LiDAR simulations. For this study, we selected the Nishishinjuku map and focused on a specific intersection for demonstration purposes. The Nishishinjuku map since it provides a challenging and realistic urban environment for evaluating LiDAR placement.

As shown in Figure~\ref{fig:roi}, we defined a ROI represented by a red cuboid with dimensions of $w = 50\,\mathrm{m}, l = 100\,\mathrm{m}, h = 5\,\mathrm{m}$. To capture traffic dynamics, we collected data every $0.25\,\mathrm{s}$ for each frame at time $t$. Within the ROI, we checked whether each voxel $v_i \in \mathcal{V}$ was occupied by objects such as vehicles, buildings, or the ground, using Unity’s simulation environment. This information was used to construct a Traffic Probabilistic Occupancy Grid (TPOG) for the ROI. The probability of a voxel $v_i$ being occupied across all $T$ frames is calculated as:

\begin{equation}
    p(v_i) = \sum_{t=1}^T \frac{\mathds{1}(\mathrm{Occ}(v_i, t))}{T}.
\end{equation}

Here, $\mathrm{Occ}(v_i, t)$ is a function that tests whether voxel $v_i$ is occupied at time $t$, and $\mathds{1}(\cdot)$ is an indicator function that outputs 1 if the voxel is occupied and 0 otherwise. For this study, we generated random traffic in AWSIM consisting of 50 vehicles and recorded data for a total of 5000 frames. Vehicles were spawned randomly within the map, following predefined road lanes and avoid collision with other vehicles. The traffic simulation accounts for realistic vehicle behaviors, such as acceleration, deceleration, and lane changes.

\begin{figure}
    \centering
    \includegraphics[width=0.95\linewidth]{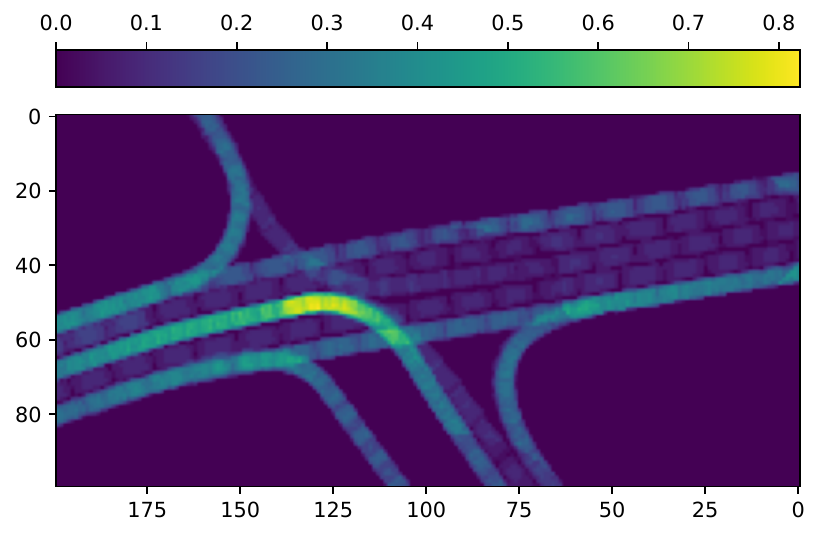}
    \caption{The traffic probabilistic occupancy grid (TPOG) from a top-down view. The occupancy probability of each grid is summed through the Z-axis for display.}
    \label{fig:pog}
\end{figure}

In Figure~\ref{fig:pog}, we visualize the TPOG from a top-down perspective by summing the occupancy probabilities along the Z-axis. The visualization reveals that traffic density is concentrated on specific lanes rather than being uniformly distributed, reflecting realistic traffic patterns.

\subsection{Surrogate Metric for Perception Performance Evaluation}

Suppose the LiDAR under consideration has $L$ scan lines and a horizontal resolution of $\psi$. In each frame, the LiDAR generates $N_B = L \cdot 2\pi / \psi$ points. To evaluate the perception performance at a specific position, we first collect point cloud data in the absence of any vehicles on the road. By connecting each point to the LiDAR’s position, we compute a set of laser segments $B$:

\begin{equation}
    B = \{b_1, b_2, \cdots, b_{N}\}.
\end{equation}

During this step, the LiDAR beams are restricted to hit the bounding box of the ROI, ensuring that all laser segments lie entirely within the ROI. Unlike vehicle-mounted LiDARs, where the local environment changes dynamically as the vehicle moves, roadside LiDARs face a static environment. If a stationary object partially blocks the LiDAR’s view, the blocked area remains unchanged over time. By collecting point cloud data in the absence of vehicles, rather than generating laser segments purely based on theoretical beam angles, our surrogate metric accounts for occlusions caused by stationary obstacles. This results in a reduced metric value when the LiDAR beams are obstructed.

Using a ray tracing algorithm, we then determine all the voxels that intersect with each laser segment $b \in B$:

\begin{equation}
    \mathcal{V}^b = \{v_1^b, v_2^b, \cdots, v_N^b\}.
\end{equation}

For a given roadside LiDAR configuration $C$, we calculate the surrogate metric Entropy-Guided Visibility Score (EGVS) $E(C)$ by measuring the information gain provided by each LiDAR beam. This is given by:

\begin{equation}\label{eq:surrogate_metric}
    E(C) = \sum_{i = 1}^M H(v_i) \cdot \min(\gamma, \sum_{j = 1}^{|B|} \mathds{1}_{v_i}(\mathcal{V}^{b_j})).
\end{equation}

Here, $H(v_i)$ represents the information value of voxel $v_i$, and $\mathds{1}_{v_i}(\cdot)$ is an indicator function that checks whether voxel $v_i$ is intersected by a given laser segment $b_j$. We want to encourage LiDAR placements where multiple beams hit voxels with high entropy to improve detection capabilities in high-traffic areas. In Equation~\ref{eq:surrogate_metric}, the term $\sum_{j = 1}^{|B|}\mathds{1}_{v_i}(\mathcal{V}^{b_j})$ represents the number of laser beams intersecting a voxel $v_i$. However, detection precision does not significantly improve once a sufficient number of beams already hit a voxel. To address this, we cap the maximum contribution from multi-beam hits using the parameter $\gamma$. The indicator function $\mathds{1}_{v_i}(\mathcal{V})$ outputs 1 if $v_i \in \mathcal{V}$ and 0 otherwise. The value of $\gamma$ should be determined based on the voxel size $\delta$ and the LiDAR model.

Since each voxel only has two states—occupied or unoccupied—we apply Shannon entropy to quantify the amount of information in each voxel. This is calculated using the occupancy probability $p(v_i)$ and its complement $p'(v_i)$, as shown below:

\begin{equation}\label{eq:entropy1}
    p'(v_i) = 1 - p(v_i),
\end{equation}
\begin{equation}\label{eq:entropy2}
    H(v_i) = -p(v_i)\log_2 p(v_i) - p'(v_i)\log_2 p'(v_i).
\end{equation}

\begin{figure}[]
\vspace{0.3cm}
    \centering
    \subfloat[A LiDAR frame recorded with no traffic inside the ROI.] {
        \includegraphics[width=0.3\textwidth,angle=90]{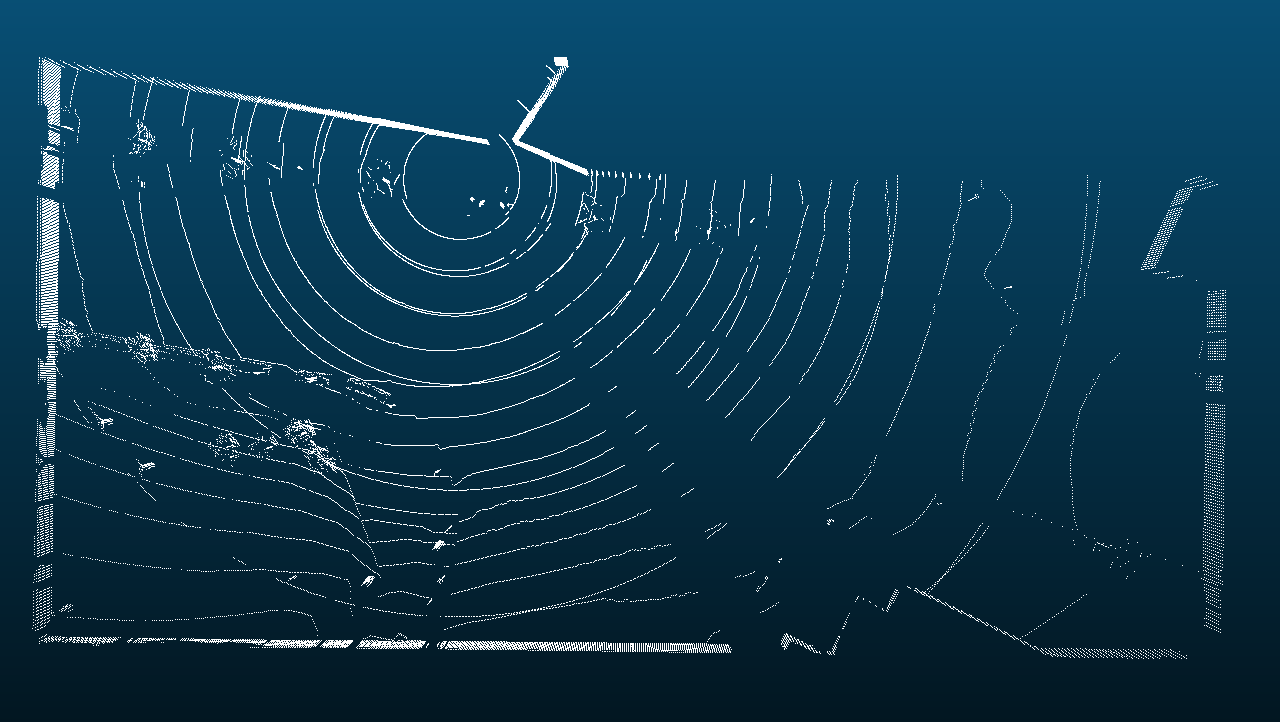}
        \label{fig:empty_frame}
    }\qquad
    \subfloat[All voxels which intersect with the LiDAR beams inside the ROI obtained using ray tracing.] {
        \includegraphics[width=0.3\textwidth,angle=90]{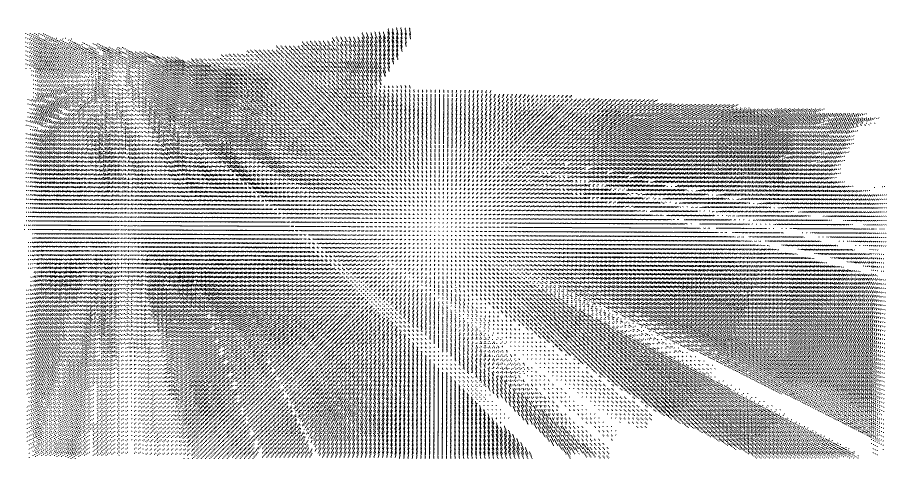}
        \label{fig:ray_tracing}
    }
    \caption{These two images demonstrate a case where occlusion reduces the surrogate score. Light-colored strips in Figure~\ref{fig:ray_tracing} represent occlusions in the LiDAR view. Entropy in these voxels does not contribute to the surrogate metric score since no laser beams intersect them, leading to reduced detection rates when vehicles occupy these voxels. A 3D view of the LiDAR is shown in Figure~\ref{fig:occlusion}.}
    \label{fig:occlusion_demo}
\end{figure}

\begin{figure}
    \centering
    \includegraphics[width=\linewidth]{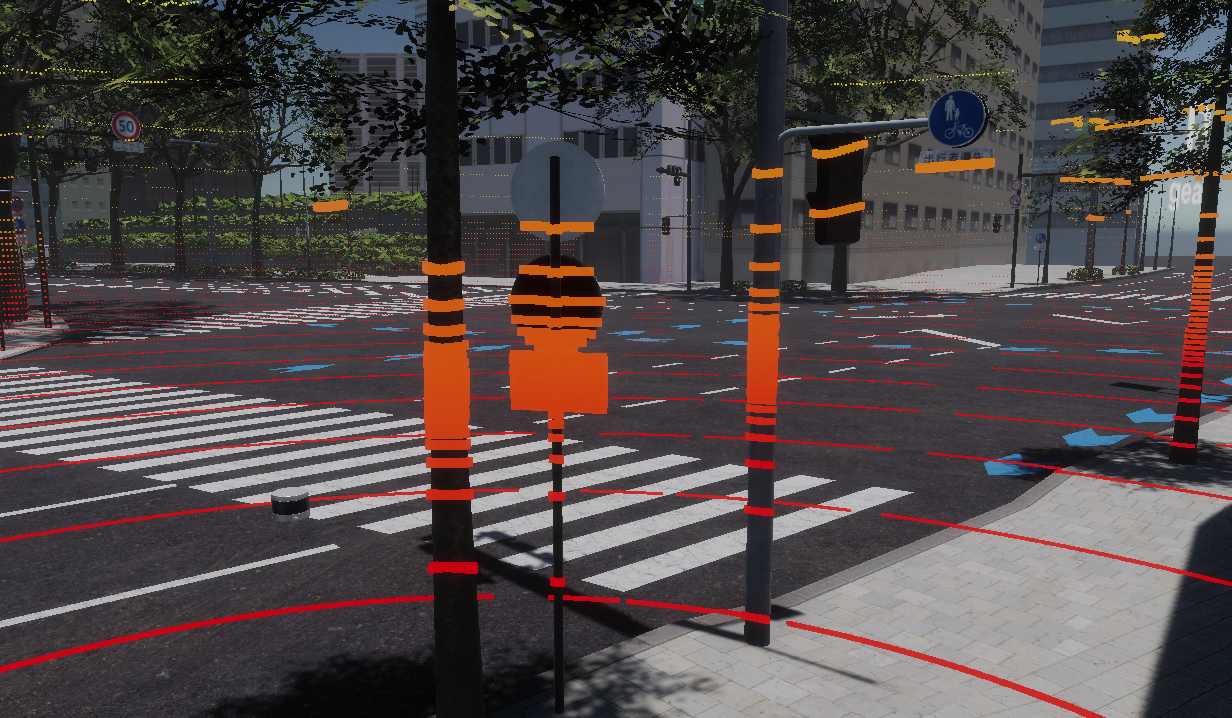}
    \caption{The view of the LiDAR is occluded by traffic lights, trees, and poles.}
    \label{fig:occlusion}
\end{figure}

Figures~\ref{fig:empty_frame} and~\ref{fig:ray_tracing} illustrate how occlusions reduce the surrogate score. In Figure~\ref{fig:ray_tracing}, light-colored strips represent areas where the LiDAR view is occluded, meaning no beams intersect these voxels. As a result, the entropy in these voxels does not contribute to the surrogate metric score, leading to a decrease in detection performance when those voxels are occupied by vehicles. A 3D representation of LiDAR occlusions caused by traffic lights, trees, and poles is shown in Figure~\ref{fig:occlusion}.

\section{Experiments and Results}
\subsection{Experiment Setup}

To validate the effectiveness of our proposed EGVS in representing the object detection ability of roadside LiDAR configurations, we conducted experiments to evaluate its performance. For these experiments, we employed PointPillars~\cite{Lang2019-ym} as the 3D object detection model. PointPillars was chosen because of its efficiency and strong performance in 3D object detection tasks, making it well-suited for roadside LiDAR evaluation.

In the experiments, AP values served as the ground truth to evaluate the reliability of EGVS. During the data collection phase, we spawned 50 vehicles in AWSIM, using a different random seed compared to the traffic pattern used for EGVS calculation. This ensures that our surrogate metric is not biased towards a single traffic pattern. To maintain consistency and fairness, however, we used the same random seed across different LiDAR configurations during this data collection phase.

We selected the VLP-32C, a 32-beam 360\degree\ LiDAR with a vertical field of view (VFoV) of +15\degree\ to -25\degree\ and a horizontal resolution of $\psi = 0.2$\degree. The simulated LiDAR operated at 10 Hz, and we collected 1000 frames of point cloud data for each configuration, along with annotations of ground truth bounding boxes. The collected data was converted into the KITTI 3D object detection format~\cite{Geiger2013-qw} to facilitate AP evaluation. The point cloud data was transformed into LiDAR-centric coordinates and cropped to match the ROI. We used the open-source OpenPCDet framework~\cite{Team-Openpcdet-Development2020-ut} to perform perception evaluation on the datasets generated for different LiDAR configurations. We compared the EGVS scores for different LiDAR configurations with their corresponding AP values, analyzing the correlation to verify the validity of EGVS as a surrogate metric. We evaluated a variety of LiDAR placements, varying parameters such as height, horizontal position, to assess the robustness of EGVS.

\begin{figure*}
    \centering
    \subfloat[EGVS versus AP values] {
        \includegraphics[width=0.5\textwidth]{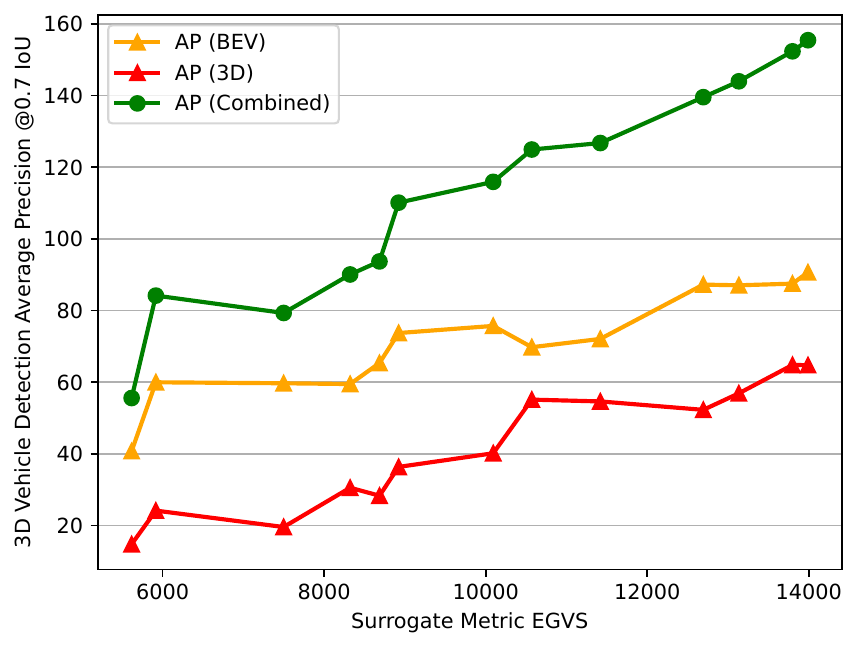}
        \label{fig:ap_plot}
    }
    \subfloat[EGVS versus AP\_R40 values] {
        \includegraphics[width=0.5\textwidth]{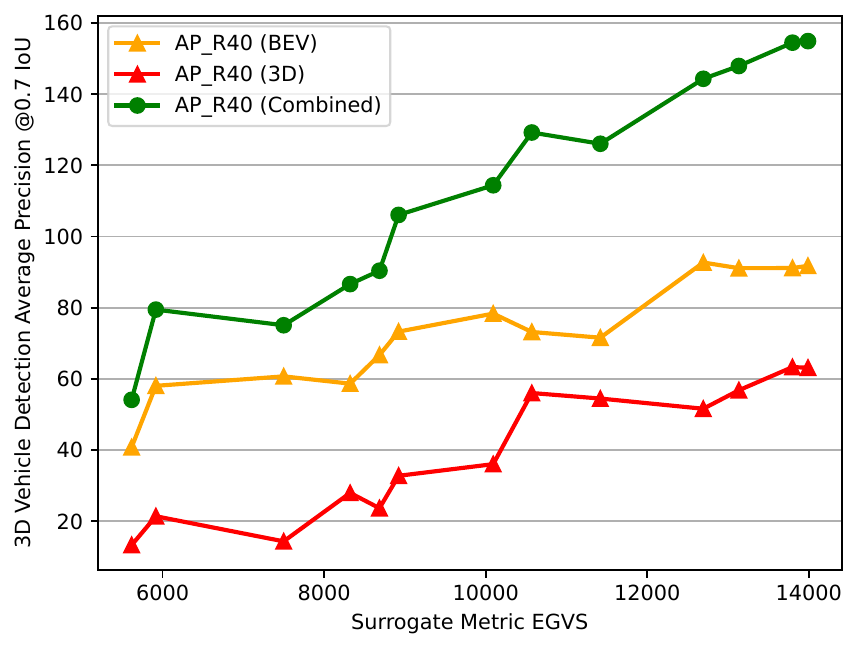}
        \label{fig:ap_r40_plot}
    }
    \caption{The relationship between the proposed surrogate metrics versus the AP of 3D vehicle detection for different LiDAR placement configuration.}
    \label{fig:results}
\end{figure*}

\subsection{Results and Discussion}
% \usepackage{booktabs}
% \usepackage{multirow}
\begin{table*}[]
\centering
\caption{Comparison of object detection performance and EGVS for different LiDAR configurations.}
\label{tab:results}
\begin{tabular}{@{}cccccccccccc@{}}
\toprule
\multicolumn{1}{c}{\multirow{2}{*}{\#}} & \multicolumn{3}{c}{LiDAR Configuration}                               & \multicolumn{1}{c}{\multirow{2}{*}{EGVS}} & \multicolumn{6}{c}{Average Precision @ 0.7 IoU}                                            \\ \cmidrule(r){2-4} \cmidrule(l){6-11}
\multicolumn{1}{c}{}   & \multicolumn{1}{c}{x} & \multicolumn{1}{c}{y} & \multicolumn{1}{c}{z} & \multicolumn{1}{c}{}                      & AP (BEV) & AP (3D) & AP (Combined) & AP\_R40 (BEV) & AP\_R40 (3D) & AP\_R40 (Combined) \\ \midrule
1                      & 195                   & 255                   & 2                     & 13990                                      & 90.66    & 64.77    & 155.43         & 91.75          & 63.14          & 154.89             \\
2                      & 200                   & 255                   & 2                     & 13798                                      & 87.51    & 64.82    & 152.33         & 91.14          & 63.31          & 154.45             \\
3                      & 190                   & 255                   & 2                     & 13134                                      & 87.05    & 56.91    & 143.96         & 91.10          & 56.80          & 147.90             \\
4                      & 185                   & 255                   & 2                     & 12694                                      & 87.24    & 52.30    & 139.54         & 92.69          & 51.61          & 144.30             \\
5                      & 220                   & 260                   & 2                     & 11419                                      & 72.09    & 54.65    & 126.74         & 71.57          & 54.48          & 126.05             \\
6                      & 175                   & 258                   & 2                     & 10093                                      & 75.72    & 40.19    & 115.91         & 78.34          & 36.05          & 114.39             \\
7                      & 225                   & 260                   & 2                     & 10571                                      & 69.78    & 55.15    & 124.93         & 73.18          & 56.03          & 129.21             \\
8                      & 175                   & 260                   & 2                     & 8922                                       & 73.73    & 36.38    & 110.11         & 73.28          & 32.78          & 106.06             \\
9                      & 165                   & 260                   & 2                     & 8685                                       & 65.36    & 28.35    & 93.71          & 66.71          & 23.68          & 90.39              \\
10                     & 200                   & 255                   & 3                     & 8322                                       & 59.50    & 30.57    & 90.07          & 58.66          & 27.97          & 86.63              \\
11                     & 220                   & 260                   & 3                     & 7499                                       & 59.72    & 19.61    & 79.33          & 60.71          & 14.36          & 75.07              \\
12                     & 175                   & 260                   & 3                     & 5917                                       & 59.99    & 24.19    & 84.18          & 58.05          & 21.41          & 79.46              \\
13                     & 150                   & 265                   & 2                     & 5617                                       & 40.78    & 14.82    & 55.60          & 40.81          & 13.31          & 54.12              \\ \bottomrule
\end{tabular}
\end{table*}

The results shown in Table~\ref{tab:results} and Figure~\ref{fig:results} demonstrate the effectiveness of the proposed surrogate metric, EGVS, in evaluating LiDAR configurations for 3D vehicle detection. The EGVS values correlate strongly with the AP metrics across both bird’s-eye view (BEV) and 3D perspectives, as well as their corresponding R40 values. The AP metric uses the 11-point interpolation method, while AP\_R40 employs 40 recall points for a more fine-grained evaluation, often leading to slightly lower values due to stricter recall granularity. The BEV metric is based on Intersection of Union (IoU) for bounding boxes from a top-down view, whereas 3D metrics are calumniating the IoU of the 3D bounding box volume. Therefore 3D metrics are inherently more difficult than BEV. As the 3D and BEV perception results can serve different purposes in roadside LiDAR applications, we combine them into a new ``combined'' metric to provide a more comprehensive measure of roadside LiDAR's detection performance.

In the experiment results, configurations with higher EGVS values consistently achieve the highest BEV AP and 3D AP scores, as well as the highest combined AP metrics. Similarly, BEV AP R40 and 3D AP R40 metrics exhibit similar trends, with higher EGVS values leading to better overall detection performance. The plots in Figure~\ref{fig:results} further illustrate the relationship between EGVS and detection performance. Both AP and AP R40 metrics show a clear upward trend as EGVS increases, especially when the EVGS value is high, which highlights the EGVS’s ability to predict detection performance reliably.

Interestingly, configurations with lower EGVS values (e.g., configuration \#13) result in significantly reduced detection performance, with combined AP score dropping to 55.60 and combined AP R40 score to 54.12. This suggests that suboptimal LiDAR placements lead to degraded data quality and insufficient spatial coverage for effective object detection. Moreover, configurations with higher $z$ values (e.g., configurations \#10 and \#11) exhibit lower performance compared to their counterparts with lower $z$, indicating that LiDAR height plays a crucial role in achieving better detection accuracy.

In summary, the proposed EGVS metric proves to be a reliable surrogate for evaluating LiDAR placement configurations, with higher values corresponding to superior detection performance in both BEV and 3D perspectives. This underscores the importance of optimizing LiDAR positioning to maximize the effectiveness of 3D object detection systems, particularly in applications like autonomous driving where precision is critical.

The evaluation of a LiDAR configuration's performance using traditional methods involves multiple steps, including simulation setup, running real-time simulations, converting data into specific formats, and using object detection models to detect vehicles and compare the results with ground truth labels. In contrast, our proposed framework significantly reduces this complexity. Using EGVS, we only need to collect traffic data within the ROI and capture a single empty LiDAR frame for each configuration. This approach also greatly reduces the time required to evaluate and compare multiple configurations. In our testing environment, which uses a Ryzen 5950X CPU with a Nvidia RTX 3090 GPU, collecting datasets and calculating AP values for a configuration typically takes several minutes, whereas it only takes serval seconds to calculate EGVS, which enables rapid evaluation and allows us to explore a larger range of possible configurations. This efficiency also makes it feasible to employ advanced methods such as gradient descent or other optimization algorithms to search for the best configuration. Additionally, EGVS benefits from computational optimizations based on the number of hitting points per voxel. When the ROI is large, both the TPOG generation and surrogate metric computation become more resource-intensive. By increasing the voxel size and compensating with the number of hitting points, we can effectively mitigate this issue and further reduce computational costs.

\section{Conclusion}
In this study, we proposed the Entropy-Guided Visibility Score (EGVS) as a surrogate metric to evaluate the object detection performance of roadside LiDAR configurations. By quantifying the information gain of LiDAR beams within a ROI and incorporating entropy to account for voxel-level information, EGVS provides a computationally efficient alternative to traditional AP-based evaluations. Unlike conventional methods that require extensive data collection, simulation, and model inference, EGVS enables rapid evaluation of LiDAR configurations using only an empty LiDAR frame and pre-collected traffic data. Through the use of EGVS, we demonstrated the potential to explore and optimize LiDAR placements more efficiently, allowing for the application of advanced search algorithms, such as gradient descent, to identify optimal configurations. The results of this work highlight the practicality and scalability of EGVS in evaluating roadside LiDAR systems, making it a valuable tool for designing and optimizing LiDAR layouts in complex urban environments.

The ultimate goal of this research is to assess the performance of roadside LiDAR placements for real-world scenarios. Future work involves exploring methods to collect traffic data using tools such as SUMO~\cite{Lopez2018-ps} combined with real-world 3D map data, like Project PLATEAU for Tokyo~\cite{Tsubaki2024-ui}, or using real-world traffic monitoring systems. This traffic data can then be converted into a TPOG to represent the traffic patterns. Real-world point cloud data will be collected for each placement to perform ray tracing and calculate EGVS. Furthermore, the proposed method can be extended to scenarios involving multiple LiDARs, where the combined coverage and AP are jointly optimized. This would enable the design of more comprehensive and robust roadside sensing systems capable of addressing the challenges of complex urban environments.

\section*{Acknowledgment}
This work was supported by JST ASPIRE Grant Number JPMJAP2325, Japan. Partial financial support was received from the Suzuki Foundation and TIER IV, Inc.

\bibliographystyle{ieeetr}
\bibliography{main}

\end{document}